\documentclass[runningheads,a4paper]{llncs}

\usepackage{amssymb}
\usepackage{graphicx}
\usepackage{xcolor}
\usepackage{url}

\urldef{\mailsa}\path|{onevzoro, nikita.zhiltsov, 
alik.kirillovich, elipachev}@gmail.com|
\newcommand{\keywords}[1]{\par\addvspace\baselineskip
\noindent\keywordname\enspace\ignorespaces#1}
\newcommand{\OntoMathPro}{$OntoMath^{PRO}$}

\begin{document}

\mainmatter  

\title{{\OntoMathPro} Ontology: \\ A Linked Data Hub for Mathematics}

\titlerunning{{\OntoMathPro} Ontology: A Linked Data Hub for Mathematics}

%
%
\author{Olga Nevzorova\inst{1,2} \and Nikita Zhiltsov\inst{1} \and Alexander Kirillovich\inst{1} \and \\ Evgeny Lipachev\inst{1}}
\authorrunning{Nevzorova et al.}

\institute{Kazan Federal University,\\
Kremlyovskaya 18 Str., 420008 Kazan, Russia\\
\mailsa\\
\and
Research Institute of Applied Semiotics of Tatarstan Academy of Sciences,\\
Baumana 20 Str., 420111 Kazan, Russia\\
}

%
%

\toctitle{\OntoMathPro Ontology: A Linked Data Hub for Mathematics}
\tocauthor{Nevzorova et al.}
\maketitle

\begin{abstract}
In this paper, we present an ontology of mathematical knowledge concepts that covers a wide range of the fields of mathematics and introduces a balanced representation between comprehensive and sensible models. We demonstrate the applications of this representation in information extraction, semantic search, and education. We argue that the ontology can be a core of future integration of math-aware data sets in the Web of Data and, therefore, provide mappings onto relevant datasets, such as DBpedia and ScienceWISE.  
\end{abstract}

\section{Introduction}\setcounter{footnote}{0}

Recent advances in computer mathematics~\cite{barendregt:challenges} have made it possible to formalize particular mathematical areas including the proofs of some remarkable results (e.g. Four Color Theorem or Kepler's Conjecture). Nevertheless, the creation of computer mathematics models is a slow process, requiring the excellent skills both in mathematics and programming. In this paper, we follow a different paradigm to mathematical knowledge representation that is based on ontology engineering and the Linked Data principles~\cite{berners:linked}. {\OntoMathPro} ontology\footnote{\url{http://ontomathpro.org}}  introduces a reasonable trade-off between plain vocabularies and highly formalized models, aiming at computable proof-checking. 

{\OntoMathPro} was first briefly presented as a part of our previous work~\cite{nevzorova:bringing}. Since then, we have elaborated the ontology structure, improved interlinking with external resources and developed new applications to support the utility of the ontology in various use cases. In summary, our key novel contributions in the current paper are:
\begin{itemize}
\item new links with external resources, such as DBpedia and ScienceWISE (Section~\ref{interlinking});
\item a concept-based mathematical formula search mashup (Section~\ref{formula:search});
\item experimental results on using the ontology in the learning process (Section~\ref{education}).
\end{itemize}

\subsection{Motivation}

The advent of the Web of Data~\cite{bizer:linked} has opened many promising technologies to publish heterogeneous data from different content providers as a single interconnected cloud of objects. We argue that the benefits of having an ontological model for mathematics and publishing mathematical knowledge as Linked Data include unification of the terminology for mathematicians, the convenient representation for applications in text mining and search, assistance in learning about mathematics, and the possibility of predicting unknown links between mathematical concepts.

\textbf{Interoperability}. Organizing scientific knowledge is utterly important for distributed teams working on large research projects. For example, it is illustrated by the emergence of ScienceWISE project~\cite{sciencewise:demo} and its ontology\footnote{\url{http://sciencewise.info/ontology/}} for physicists in CERN. Since {\OntoMathPro} has better coverage than Wikipedia regarding the developing vocabulary and particularly object properties, it can serve as the main repository for definitions of mathematical concepts in the Web of Data. It means that mathematicians may unambiguously refer to the ontology concepts via URIs on discussion groups, blogs, and trendy Q\&A sites, such as MathOverflow\footnote{\url{http://mathoverflow.net/}}. For this purpose, we provide a URI lookup service as well as a URI dereferencing service (Section~\ref{concepts}).

\textbf{Convenient format for mashups}. The integrated representation of mathematical knowledge in a machine-readable format (RDF) may boost the development of new handy services, i.e., Semantic Web agents, for mathematicians. Such services could be run atop the ontology as well as datasets, modeled with the help of the ontology. In Section~\ref{applications}, we present our demo applications in text mining and mathematical formula search, which exploit the ontology as a rich linguistic resource.

\textbf{Learning}. From the learner's perspective, the ontology gives the helpful context for conceiving a mathematical term, including the definition, related concepts with respect to non-trivial relations, such as logical dependency and association. Besides, we argue that the ontology can facilitate educational assessment of students. In Section~\ref{education}, we describe our experiments on using {\OntoMathPro} as a tool for measuring the effectiveness of the course on numerical analysis.

\textbf{Discovering hidden links}. The ontological model generally defines not only concepts from the domain of interest, but also relations between them and axioms (e.g. transitivity or cardinality of relations). Thus, ontologies may enable inference over knowledge bases of facts. {\OntoMathPro} has a rich set of relations between mathematical concepts. We expect that existing link prediction techniques (e.g.~\cite{bordes:translating,nickel:rescal,sutskever:bctf}) along with reasoning mechanisms (e.g.~\cite{pellet}) may reveal compelling hidden relationships between known concepts in mathematics. Such discovered highly probable relations may guide further research in the bleeding edge of mathematics, highlighting the most prospective directions.

\subsection{Related Work}

To put our research into the context, we summarize the most relevant previous works for representing mathematical knowledge in this section. For a more comprehensive overview of services, ontological models and languages for mathematical knowledge management on the Semantic Web and beyond, we refer the interested reader to C. Lange's survey~\cite{lange:review}.

\textbf{Symbolic notation}. The semantic layer of Mathematical Markup Language (MathML)~\cite{mathml,elizarov:mathml} -- Content MathML -- as well as OpenMath Content Dictionaries~\cite{openmath} are extensible collections of definitions of symbols. Basically, they suffice high school and sophomore level education: arithmetics, set theory, calculus, algebra, etc. Each symbol has its own URI. In comparison, {\OntoMathPro} does not contain definitions of symbols and could be easily integrated with Content MathML/OpenMath dictionaries. 

\textbf{High-level ontologies}. 
Next, we overview ontologies for representing high-level structures in the mathematical knowledge: OMDoc~\cite{kohlhase:omdoc,lange:thesis}, MathLang's Document Rhetorical aspect (DRa) Ontology~\cite{kamareddine:mathlang}, Mocassin Ontology~\cite{mocassin}. These models enable making closely related assertions for the particular fields of mathematics, i.e., theories. Comparing to them, {\OntoMathPro} rather specifies theories themselves.

Open Mathematical Documents (OMDoc), an XML-based language, is integrated with MathML/OpenMath and adds support of statements, theories, and rhetorical structures to formalize mathematical documents. OMDoc has been used for interaction between structured specification systems and automated theorem provers. The OMDoc OWL Ontology\footnote{Available at \url{http://kwarc.info/projects/docOnto/omdoc.html}}
is based on the notion of statements. Sub-statement structures include definitions, theorems, lemmas, corollaries, proof steps. The relation set comprises of partonomic (whole-part), logical dependency, and verbalizing properties. The paper~\cite{david} presents an OMDoc-based approach to author mathematical lecture notes and expose them as Linked Data.

The MathLang DRa Ontology characterizes document structure elements according to their mathematical rhetorical roles that are similar to the ones defined in the statement level of OMDoc. This semantics focuses on formalizing proof skeletons for generation of proof checker templates.

The Mocassin Ontology encompasses many structural elements of the state-of-the-art models. However, this model is more oriented on representing structural elements that occur in real scholarly papers on mathematics. Our previous work~\cite{mocassin} demonstrates its utility in the information extraction scenario.

\textbf{Terminological vocabularies}.
The general-purpose DBpedia dataset~\cite{dbpedia} contains, according to our estimates, about 7,800 concepts (including 1,500 concepts with labels in Russian)  from algebra, 46,000 (9,200) concepts from geometry, 30,000 (4,300) concepts from mathematical logic, 150,000 (28,000) from mathematical analysis, and 165,000 (39,000) concepts on theory of probability and statistics. Concepts are linked to DBpedia categories representing the fields of mathematics. Although there is a \emph{skos:broader} relation between categories, there is no taxonomic (ISA) relationship between the concepts themselves.

A SKOS-based adaptation of Mathematics Subject Classification\footnote{\url{http://www.ams.org/msc/‎}} is exposed as a linked dataset~\cite{msc:eswc2012}. {\OntoMathPro} ontology overlaps with this dataset in case of modeling hierarchy of fields, but it is significantly richer for representing terms and their interactions.

The Online Encyclopedia of Integer Sequences~\cite{oeis} is a knowledge base of facts about numbers. Given a sequence of integers, this service\footnote{\url{http://oeis.org}} returns the information about its name, general formula, implementation in programming languages, successive numbers, references, and other relevant links.

\textbf{Thesauri and ontologies}. Hence, let us consider domain-specific resources, providing a more rich set of relations.
 \cite{gruber} presents a formal ontology of mathematics for engineers that covers abstract algebra and metrology. Cambridge Mathematical Thesaurus~\cite{cmt} contains a taxonomy of about 4,500 entities in 9 languages from the undergraduate level mathematics, connected with logical dependency \emph{referencedBy} and associative relationships \emph{seeAlso}. This resource has been developed in education purposes and covers only bachelor level mathematics. 
 
 The ScienceWISE project ontology~\cite{sciencewise:demo} gives over 2,500 mathematical definitions connected with ISA-, whole-part, associative, and importance relationships. The sources of definitions are Wikipedia, Encyclopedia of Science, and the engaged research community. The project focuses on achieving a consensus of opinion among mathematicians about given definitions. 

The Ontology on Natural Sciences and Technology~\cite{dobrov:oent} contains 55,000 descriptions of scientific terms in Russian, covering the mathematical terminology on high school and freshman-sophomore university levels. The ontology is meant for applications of text analysis, and defines thesaurus-like relations, such as ISA, whole-part, asymmetric association, and symmetric association.

Due to the lack of space, we do not cover related works on semantic data analysis for mathematical texts, which are given in~\cite{ntis,mtc}. 

\section{{\OntoMathPro} Structure}

In this section, we elaborate the modeling principles, the development workflow, the ontology structure, and links to external terminological resources.

\subsection{Modeling Principles}

Even though mathematics is the most exact science, modeling this domain is hard, due to:
\begin{itemize}
\item \emph{abstractness}, i.e., many definitions are conventionally given in mathematical notation elements or formulas;
\item \emph{duality}, i.e., there might be equivalent definitions for terms depending on which foundations of mathematics are used (set theory or geometry) -- this aggravates asserting logical dependency relations between concepts;
\item \emph{emergence of novel terms}, i.e., developing, not commonly used parts of the vocabulary in the professional community.

\end{itemize}

To tackle these issues, we come up with the following  modeling principles.

\begin{enumerate}
\item \textbf{Only classes, no individuals}. First, {\OntoMathPro} is geared to be a linguistic resource for text processing. Therefore, the ontology does not contain individuals. The latter can be found in applications, e.g. concrete occurrences of named entities in texts. For example, while modeling mathematical numbers like $\pi$ or $e$ as individuals is natural, we model them as classes, because, in our case, individuals can be occurrences of these numbers in texts.

\item \textbf{ISA vs. whole-part}. Existing classification schemes, such as MSC or UDC\footnote{\url{http://www.udcc.org/}}, model hierarchies with respect to whole-part relation.
Unlike them, our ontology posits the ISA semantics for hierarchies of mathematical knowledge objects, and preserves the same for fields and sub-fields. The reason is that there are only classes instead of individuals in {\OntoMathPro}, we express the whole-part semantics through ISA relation taking into account its interpretability in terms of the set theory. Thus, we assume that a field of mathematics is a set of closely related statements. For example, fractal geometry is a sub-set of geometry.

\item \textbf{Validating classes}. We deal with the developing vocabulary. To avoid coining a rare terminology, we require a reference from a refereed publication (i.e., an article or a textbook) for a term to be added to the ontology. 

\item \textbf{Validating relations}. Establishing correct relation instances is hard and requires high-level competence. Therefore, we basically rely on the opinions of experienced experts involved in the development. The exceptions include the logical dependency and \emph{solves} relations that can be validated using references to refereed sources (see our explanation in Section~\ref{section:relations}).

\item \textbf{URI naming convention}. Since the ontology is bilingual (Russian/English) and our experts had started adding terms with Russian labels and translated them to English afterwards, we choose using surrogate URIs, e.g. \url{http://ontomathpro.org/ontology/E1} for a concept ``Field of mathematics"\footnote{the similar convention was adopted in CIDOC CRM Ontology~\cite{cidoc}}.
\item \textbf{Multiple inheritance}. Multiple inheritance with respect to ISA-relationships is permitted. For example, class E1892 Differential Equation is a sub-class of both E1891 Equation and E2688 Element of Differential Equations. 
\item \textbf{Synset as labels}. Synonyms are represented by labels of the same class. For example, E1226 Cauchy's Inequality has labels ``Cauchy's inequality" and ``Inequality of arithmetic and geometric means".
\end{enumerate}

We have worked with seven practicing mathematicians as domain experts for four months. The terminological sources include freely available materials, such as classical textbooks, Wikipedia articles, and real scholarly papers, along with personal experience of the experts. During the development, we used a collaborative tool WebProtege\footnote{\url{http://webprotege.stanford.edu/}}~\cite{webprotege}.

\subsection{Concepts}\label{concepts}

Each concept is represented as an OWL class in the ontology. Each class has a textual explanation or hyperlink to its external definition, Russian and English labels. All the metadata information, including adjacent properties, per each class can be seen on our URI dereferencing service\footnote{according to the Linked Data principles, it is available on the ontology's URI: \\ \url{http://ontomathpro.org/ontology}}.
To facilitate finding the URI for a given class (e.g. for adding a link to it on a webpage), this service features a URI lookup function. In future, we are going to add collecting backlinks per each class from HTTP referrers. It will allow us to display relevant webpages and discussions about the concept.

\begin{figure}
\centering
\includegraphics[height=6cm]{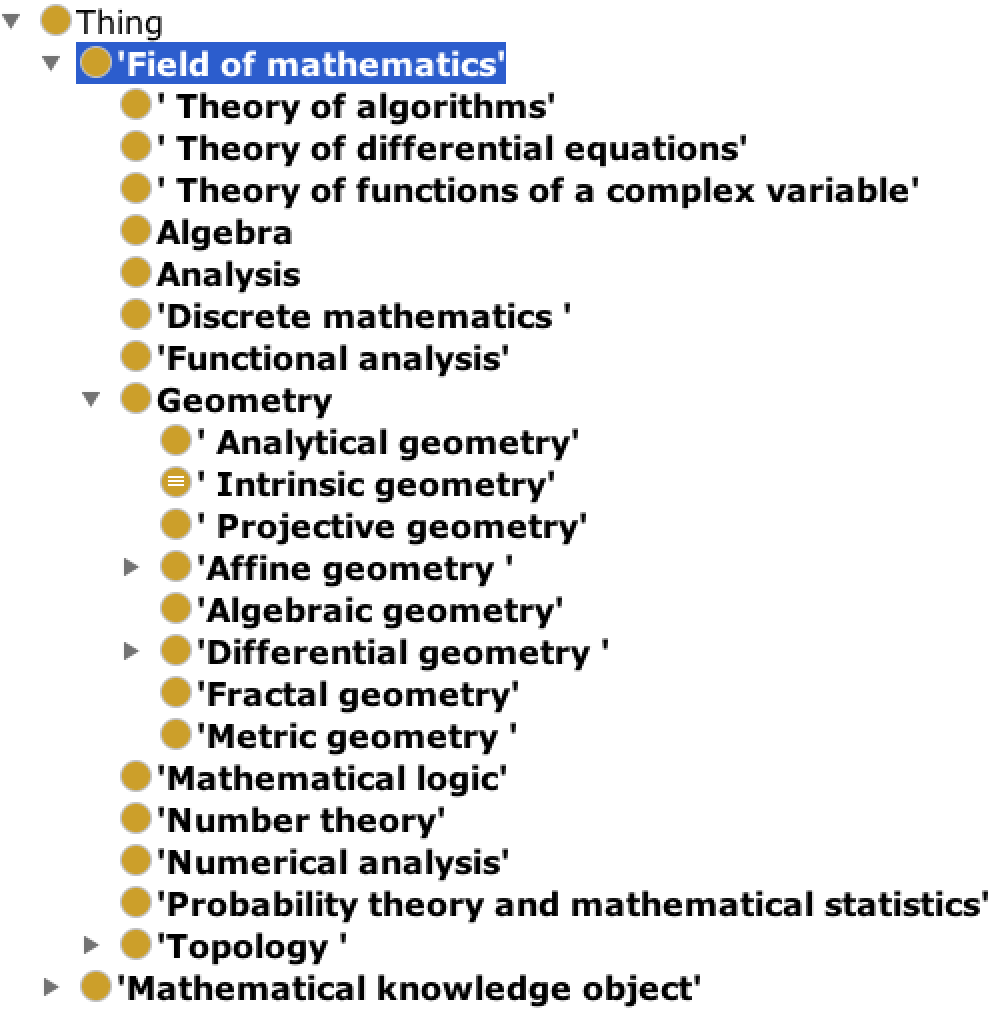}
\caption{A taxonomy of the fields of mathematics in {\OntoMathPro}}
\label{fig:fields}
\end{figure}

\begin{figure}
\centering
\includegraphics[height=6cm]{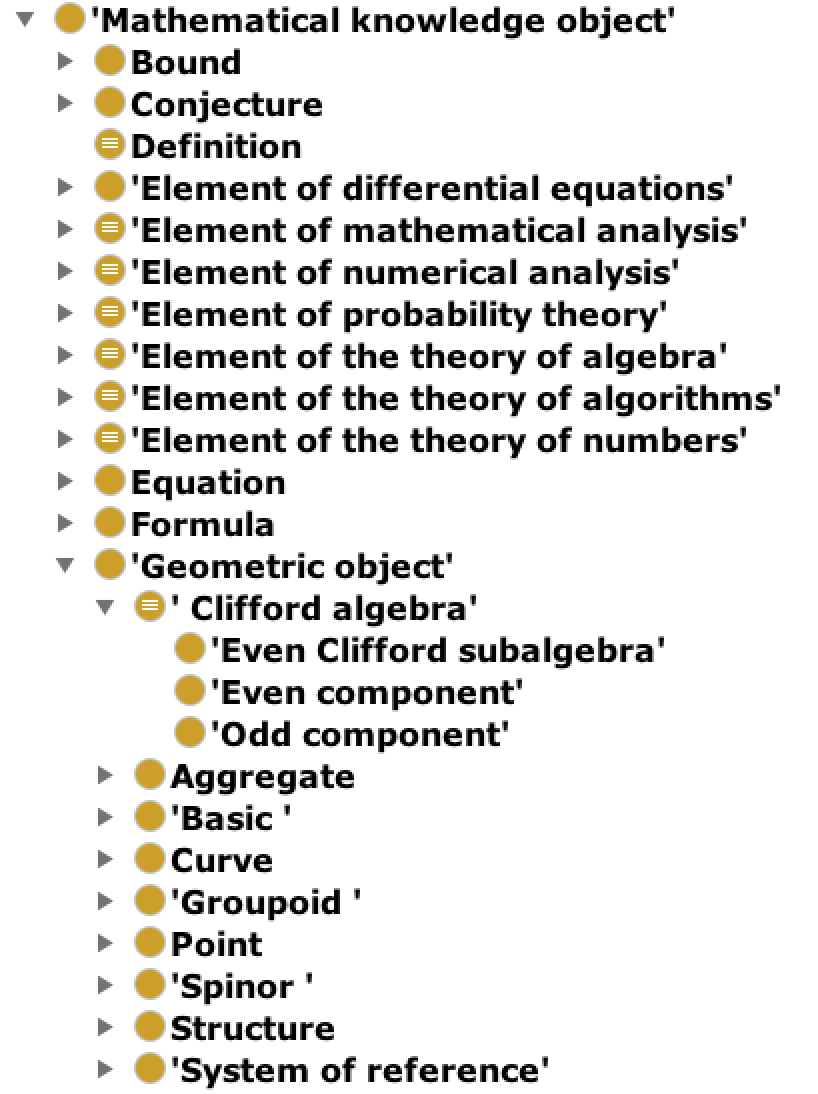}
\caption{A taxonomy of mathematical knowledge objects in {\OntoMathPro}}
\label{fig:objects}
\end{figure}

We distinguish two hierarchies of classes with respect to ISA-relationship: a taxonomy of the fields of mathematics (Figure~\ref{fig:fields}) and a taxonomy of mathematical knowledge objects, i.e., elements of particular theories (Figure~\ref{fig:objects}). In the taxonomy of fields, most fundamental fields, such as geometry and analysis, have been elaborated thoroughly. For example, there have been defined specific sub-fields of geometry: analytic geometry, differential geometry, fractal geomentry and others. The ontology covers a wide range of the fields of mathematics, such as number theory, set theory, algebra, analysis, geometry, mathematical logic, discrete mathematics, theory of computation, differential equations, numerical analysis, probability theory, and statistics.

 There are three types of top level concepts in the taxonomy of mathematical knowledge objects: i) basic metamathematical concepts, e.g. E847 Set, E1227 Operator, E1324 Map, etc; ii) root elements of the concepts related to the particular fields of mathematics, e.g. E2406 Element of Probability Theory or E3140 Element of Numerical Analysis; iii) common scientific concepts: E339 Problem, E449 Method, E1936 Statement, E1988 Formula, etc. Most concepts are inherited from the root elements of the fields of mathematics (type ii).
 
In total, {\OntoMathPro} contains 3,449 classes.

\subsection{Relations}\label{section:relations}

{\OntoMathPro} defines four types of object properties:
\begin{itemize}
\item a directed relation between a mathematical knowledge object (E24) and a field of mathematics (E1) (\emph{P3 belongs to} and its inverse \emph{P4 contains}), e.g. E68 Barycentric Coordinates \emph{P3 belongs to} E14 Metric Geometry;
\item a directed relation of logical dependency between mathematical knowledge objects (\emph{P1 defines} and its inverse \emph{P2 is defined by}), e.g. E39 Christoffel Symbol \emph{P2 is defined by} E213 Connectedness. This relation can be validated with providing a reference to the relevant definition in a refereed publication. The \emph{P2 is defined by} relation instance is established, if the definition of the first concept (``subject" in the triple)  explicitly refers to the second concept (``object"); 
\item a symmetric transitive associative relation between mathematical knowledge objects  (\emph{P5 see also}), e.g. E660 Chebyshev Iterative Method \emph{P5 see also} E444 Numerical Solution of Linear Equation Systems. The semantics of this relation is equivalent to \emph{rdfs:seeAlso}, which is widely used in the LOD datasets for individuals;
\item a directed relation between a method and a task (\emph{P6 solves} and its inverse \emph{P7 is solved by}), these properties are defined on E449 Method and E339 Problem classes. This relation can also be validated using references to refereed articles or textbooks.
\end{itemize}

In total, {\OntoMathPro} contains 3,627 ISA property instances, and 1,139 other property instances.

\subsection{Links to Other Datasets}\label{interlinking}

Although external links are no part of the ontology, we describe the subsets that have been mapped onto existing Linked Data resources.
An alignment of {\OntoMathPro} with DBpedia\footnote{SPARQL endpoint: \url{http://dbpedia.org/sparql}} is based on the following features:
\begin{itemize}
\item class and resource labels (\emph{rdfs:label} property);
\item explicit links to Wikipedia, i.e., during the development of the ontology, some definitions were imported from Wikipedia and refer to it. We compare these references with \emph{foaf:primaryTopic} and \emph{rdfs:labels} property values in DBpedia.
\end{itemize}

For interlinking, we only use DBpedia resources that belong to the Mathematics category and its subcategories (e.g. Algebra, Geometry, Mathematical logic, Dynamical Systems) up to 5 levels with respect to \emph{skos:broader} property. After the linking has been accomplished, we generate triples connecting the classes of the {\OntoMathPro} with the resources from DBpedia by using \emph{skos:closeMatch} property. The alignment with DBpedia has resulted in 947 connections with 907 the ontology classes (some classes were linked with several DBpedia resources). 

An alignment with ScienceWISE Ontology\footnote{\url{http://data.sciencewise.info/openrdf-sesame/repositories/SW}} only used class labels. As a result, this mapping comprises 347 connections.  The  mapping files are available on GitHub\footnote{\url{https://github.com/CLLKazan/OntoMathPro}}.

\section{Applications}\label{applications}

In this section, we describe the applications, which exploit different aspects of {\OntoMathPro}.

\subsection{Information Extraction}\label{extraction}

In our previous work~\cite{nevzorova:bringing}, we have developed a semantic publishing platform for scientific collections in mathematics that analyzes the underlying semantics in mathematical scholarly papers  and effectively builds their consolidated ontology-based representation. 

Thus, every paper in the collection is dissected into a semantic graph of instances of the supported domain models with the help of NLP techniques, such as noun phrase recognition and entity resolution. Along with textual fragments, the platform is capable to understand the meanings of mathematical notation symbols and interpret them as ontology instances.

The corpus of publications we experimented with consists of 1,330 articles of the ``Proceedings of Higher Education Institutions: Mathematics" (PHEIM) journal in Russian.  
The current RDF dataset\footnote{is exposed via our SPARQL endpoint: \url{http://cll.niimm.ksu.ru:8890/sparql}, the graph IRI is \url{http://cll.niimm.ksu.ru/pheim}} -- PHEIM dataset -- contains more than 850,000 triples including the descriptions of 1,330 articles, 17,397 formulas, 43,963 variables, and 66,434 textual occurrences of mathematical entities from the ontology.

\subsection{Concept-based Formula Search}\label{formula:search}

\textbf{Approach}. Our demo application\footnote{\url{http://cll.niimm.ksu.ru/mathsearch}} supports a use case of searching mathematical formulas in the published PHEIM dataset that are relevant to a given  entity. The user input supported by the application is close to a keyword search: our system is agnostic to a particular symbolic notation used in the papers to denote mathematical concepts, and the user is able to filter query suggestions by keywords. This feature makes our application different from a wide range of mathematical retrieval systems that require the input in the {\LaTeX} syntax (e.g.~\cite{springer:latex,wiki:formula,wolfram:formula}). Such approaches usually suffer from ambiguous mathematical notations. However, it's worth mentioning \emph{(uni)quation}~\cite{uniquation}, a formula search system that is robust to basic formula transformations including changes of variables. 

Another rationale behind the concept-based search input interface is its cross-language capabilities, i.e., even though the indexed document collection is in Russian, the user still can search using keywords in English.

There have been a few efforts to enable a keyword search for retrieving mathematical content. A computational engine WolframAlpha~\cite{wolfram:alpha} can handle keyword queries, generating a summary for a mathematical concept. However, the engine does not provide the similar functionality for documents in scientific collections. MCAT Math Retrieval System~\cite{mcat} applies an SVM classifier to detect descriptions of mathematical expressions and extends the TF-IDF ranking baseline for formula search in MathML documents. Unlike this tool, our solution is more powerful, because it resolves the lexical meanings of symbols in terms of {\OntoMathPro} ontology, and, therefore, enabling reasoning with respect to the ontology relations (e.g. ISA).
Additionally, our search interface supports filtering by the document structure context, i.e., a particular segment of the document (e.g. a theorem or a definition) that contains the relevant formula.

\begin{figure}
\centering
\framebox{\includegraphics[height=4.2cm]{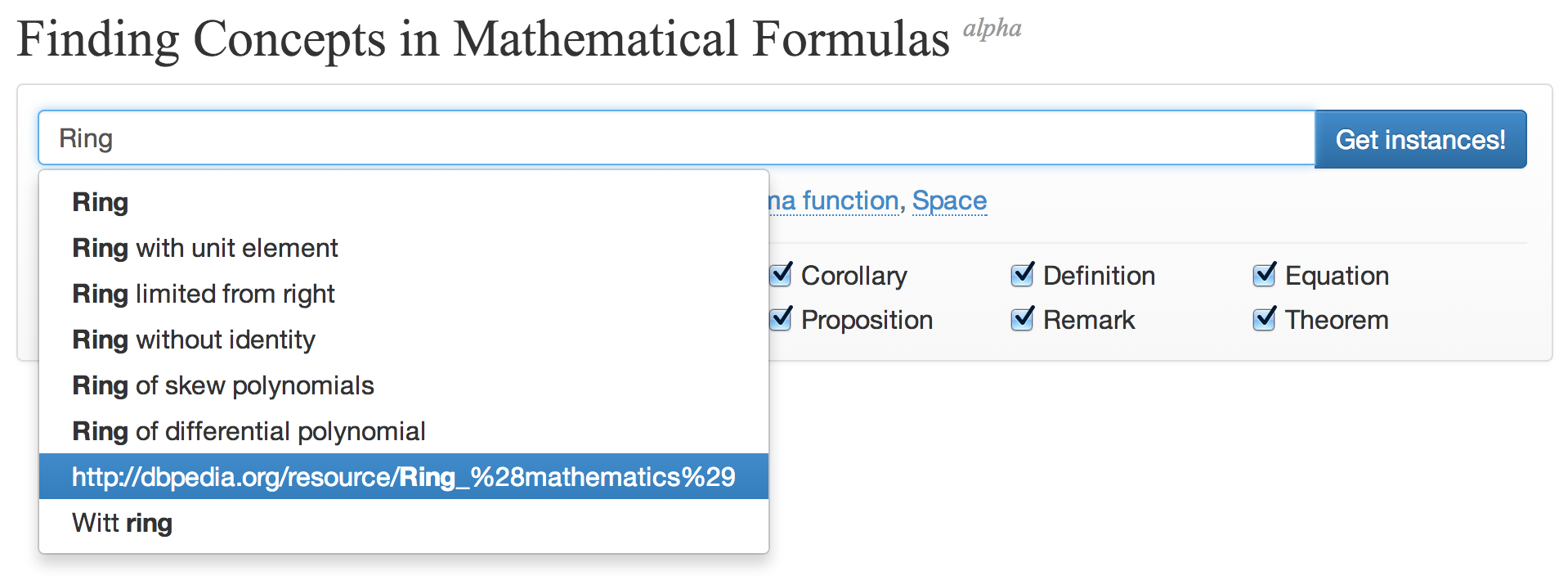}}
\caption{Query interface of the mathematical formula search demo application}
\label{fig:query}
\end{figure}

\textbf{Demo}. The application supplies the following search scenario. The user enters keywords in the search box (Figure~\ref{fig:query}) and may choose a related  concept suggestion. The suggestions are loaded from \OntoMathPro ontology as well as the DBpedia part aligned with the ontology.

\begin{figure}
\centering
\framebox{\includegraphics[height=6cm]{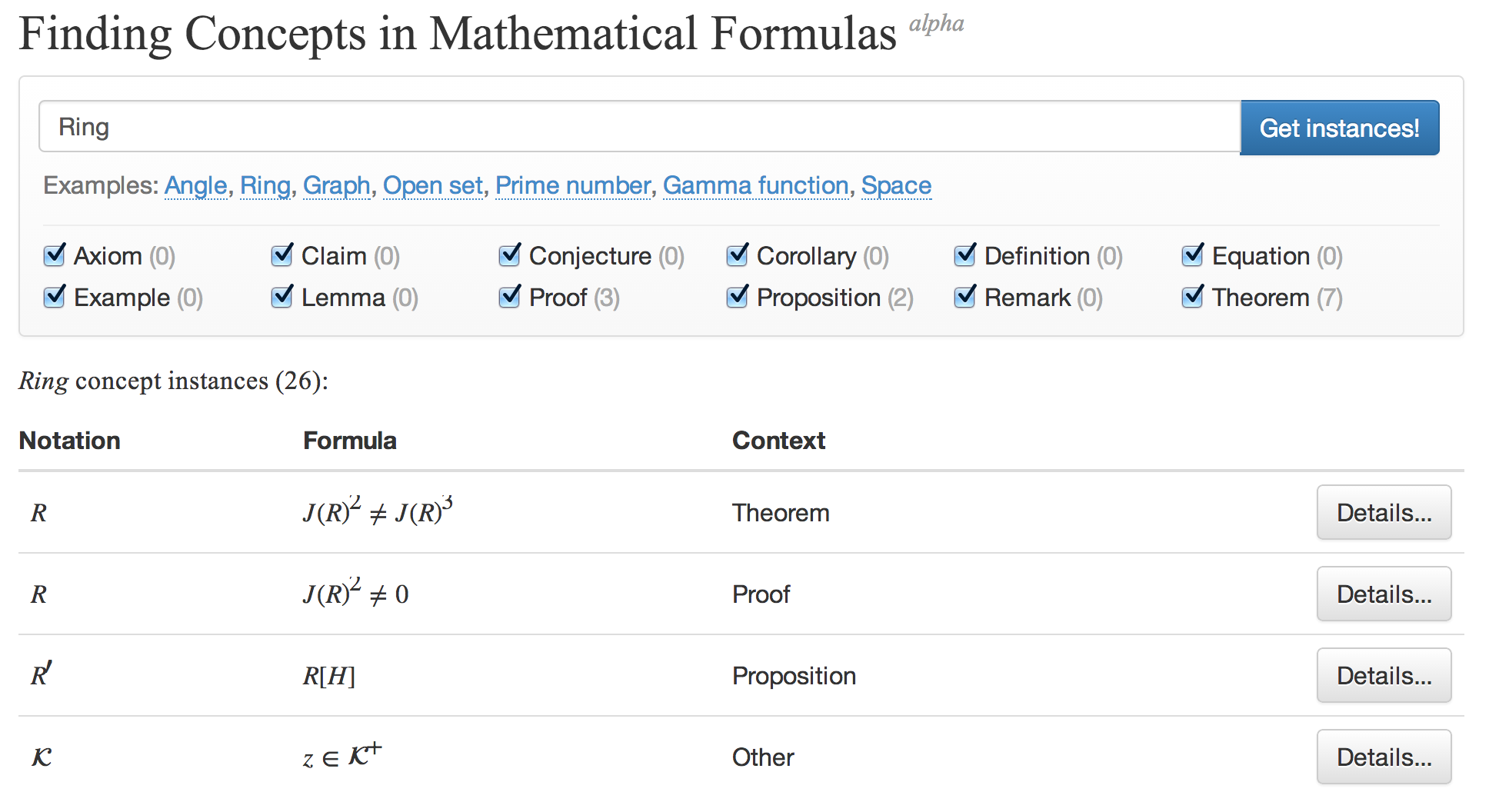}}
\caption{A search result page}
\label{fig:result}
\end{figure}

Then, the user clicks the ``Get instances" button and gets a list of search results (Figure~\ref{fig:result}). The hit descriptions are represented as table rows with the contextual information. Specifically, the application provides a symbolic interpretation of the chosen concept, the relevant formula, in which the notation element occurs, and the context, i.e., the document part, which contains the formula. Providing notation examples is instructive, because the same concept may be expressed in different documents using different symbols depending on the authors' writing style. Additionally, the user can refine the context of search results selecting proper checkboxes standing for particular document parts.

\begin{figure}
\centering
\framebox{\includegraphics[height=4cm]{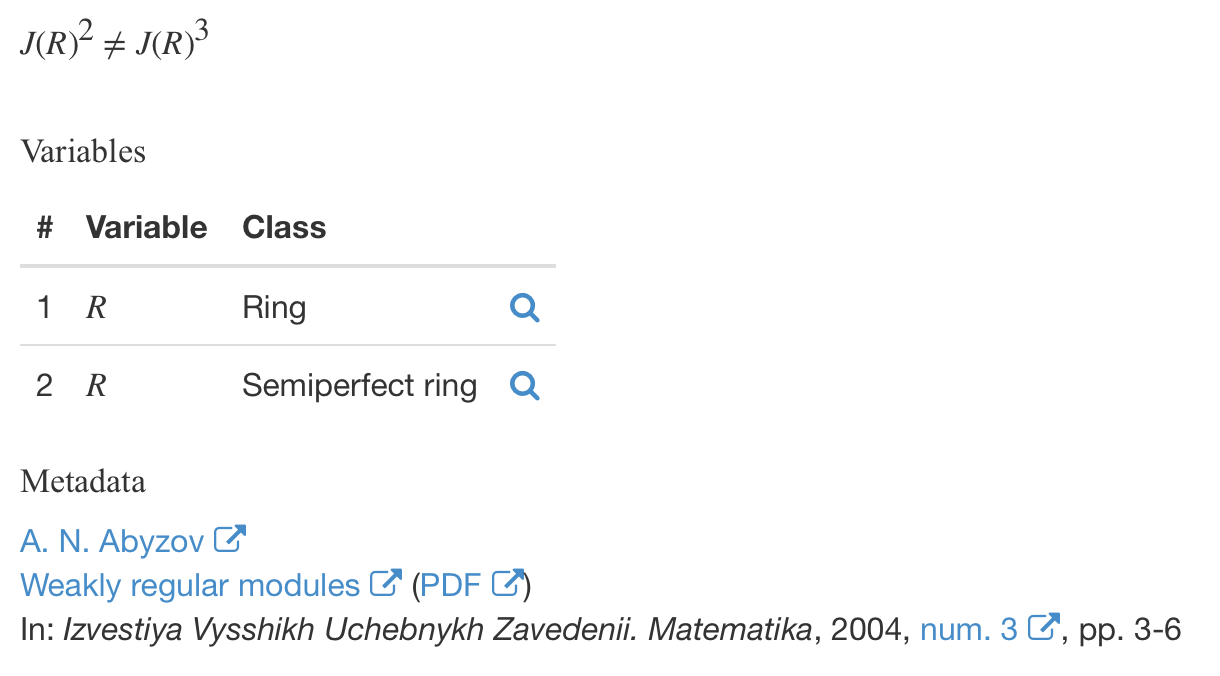}}
\caption{A search result details page}
\label{fig:details}
\end{figure}

Finally, the user can examine the paper, which contains the relevant formula after clicking on the ``Details" button (Figure~\ref{fig:details}). This screen provides the relevant formula, the matched concepts as well as the related article information including links to its metadata page and PDF.

We would like to note that the current search result set does not cover all the concepts defined in {\OntoMathPro}  ontology. The reason is twofold: i) the data set is quite limited; ii) many named entities are usually not expressed in a symbolic notation (e.g. theorems or the fields of mathematics).

From the engineering perspective, our demo application is a lightweight mashup that consumes the published RDF data set. It is written in JavaScript and queries our SPARQL endpoint.

\subsection{Education}\label{education}

In this subsection, we describe our experiments on ontology-based assessment of the competence of students, who attended a course on numerical analysis instructed by one of the authors for 3-year undergraduate students at our university.

Most conventional approaches to assessment of students still tend to evaluate how much a student knows about the subject. However, problem-based learning is proven to be a more effective approach~\cite{barrows}. For a practicing mathematician, an ability of finding a method for solving a concrete task is crucial. The proficient solver must realize relationships between particular methods, tasks and other mathematical concepts, i.e., ``know how to conceive a holistic image of the discipline"\footnote{translated from the Russian Federal Standard for Higher Education on mathematics for master students, Section 5.2: \url{http://www.edu.ru/db-mon/mo/Data/d_10/prm40-1.pdf}}. We propose to consider an ontology as a close approximation to such an image.

\begin{figure}
\centering
\includegraphics[scale=0.4]{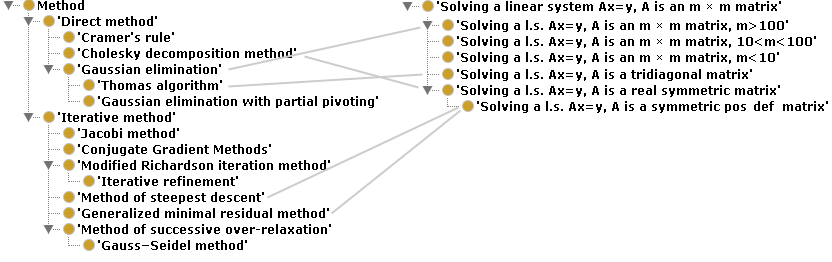}
\caption{A part of the correct solution for the interlinking task, the arrows mean \emph{P6 solves} relation instances}
\label{fig:experiment}
\end{figure}

For our experiments, we extracted a small fragment of {\OntoMathPro} ontology (Figure~\ref{fig:experiment}). It contains taxonomies of tasks and solving methods for systems of linear equations (numerical analysis) as well as relationships between them. We added a few relevant concepts that were not defined in the ontology due to questionable definitions, e.g. ``a task of solving a system of linear equations with the coefficient matrix of order m*m, case $m > 100$". However, such concepts are useful for testing pragmatic competence. 

Concerning the course coverage, only a few methods from the fragment were out of its scope. There is an assumption that students may have studied them during their research work. 

The experiment participants are 25 students who attended the course and had high overall grades. They are 3-year undergraduate students (who have not yet passed the exam on the course), 1- and 2-year master students, and Ph.D. students. Each participant is given a list of classes and asked to link them using only two relationships (ISA and \emph{P6 solves}).

Then, we stated a hypothesis that the participants' results have to correlate with their experience. We use standard performance measures for classification tasks, such as precision (P), recall (R), and F-score $=2\cdot\frac{P*R}{P+R}$. 

We emphasize the importance of applying a reasoning mechanism during the calculation of experimental results.
For the evaluation, we materialize all facts that might be acquired after reasoning. For example, a relationship \emph{P6 solves} is inherited through both the taxonomies of methods and task. Thus, if the participant links correctly more specific classes, it results in lower recall, but does not affect precision. 

\begin{table}
\center
\caption{Average results of interlinking ontology concepts from a course on numerical analysis for different groups of students. P means precision, R -- recall, F -- F-score.}
\begin{tabular}{|l|c|c|c|c|c|c|c|c|c|c|c|c|}
\hline
\textbf{Group of students} & \multicolumn{3}{c|}{\textbf{ISA Tasks}} & \multicolumn{3}{c|}{\textbf{ISA Methods}} & \multicolumn{3}{c|}{\textbf{P6 solves}} & \multicolumn{3}{c|}{\textbf{Total}} \\
\hline
& P & R & F & P & R & F & P & R & F & P & R & F \\
\hline
~3-year undergraduates~& 1.00 & 1.00 & 1.00 & 0.76 & 0.59 & 0.67 & 0.46 & 0.27 & 0.34 & 0.69 & 0.50 & 0.58 \\
~1-year masters~ & 1.00 & 1.00 & 1.00 & 0.79 & \textbf{0.63} & 0.70 & 0.43 & 0.26 & 0.32 & 0.70 & \textbf{0.52} & 0.59 \\
~2-year masters~ & 1.00 & 1.00 & 1.00 & 0.65 & 0.55 & 0.59 & 0.51 & \textbf{0.37} & \textbf{0.41} & 0.63 & \textbf{0.52} & 0.56 \\
~Ph.D. candidates~ & 1.00 & 1.00 & 1.00 & \textbf{0.86} & \textbf{0.63} & \textbf{0.71} & \textbf{0.61} & 0.23 & 0.33 & \textbf{0.82} & 0.50 & \textbf{0.62}\\
\hline
\end{tabular}
\label{student:evaluation}
\end{table}

Table~\ref{student:evaluation} shows the results of the interlinking done by the participants. As expected, establishing links in the simplistic taxonomy of tasks have been done well by all participants. The reconstruction of a slightly complex taxonomy of methods has significantly lower performance values and correlates with the background of participants. Again as expected, classification of \emph{P6 solves} relations turns out to be the hardest part. Interestingly, the performance values for this subtask correlates less with the level of participants: Ph.D. students give less complete results, than not so experienced participants. A possible explanation comes from the fact that Ph.D. students have chosen their specific directions of research, which might not overlap with systems of linear equations as a sub-field of numerical analysis. That's why they classify tasks and methods very well, but prefer to keep in mind relationships about solution methods that are only relevant to their area of expertise.

Finally, we may conclude that the proposed approach could be used for evaluation of the competence of students.

\section{Conclusion}

We present {\OntoMathPro}, an OWL ontology that is geared to be the hub of mathematical knowledge in the Web of Data. Relying on this representation model, we are going to create an ecosystem of datasets and mashups around the ontology to benefit mathematicians from different backgrounds -- from undergraduate students to experts. In this paper, we describe the first steps towards it. In particular, interlinking with other Linked Data datasets, the applications in information extraction, semantic search of mathematical formulas, learning are discussed.

We emphasize that although the ontology has achieved maturity, it is the result of ongoing work. We share the sources with the Semantic Web community and organize the further collaborative development of the ontology and its applications on the project page \url{ontomathpro.org} to engage our colleagues and mathematicians from elsewhere.

As a future research work, we plan to study topic fusion in educational mathematical documents, such as lecture notes and textbooks, through occurrences of concepts from different taxonomies of the ontology. Additionally, we will address applications of ontology reconstruction in learning purposes (as described in Section~\ref{education}), and  are going to extend this approach on other fields of knowledge.

\subsubsection*{Acknowledgments.} This work was funded by the subsidy allocated to Kazan Federal University for the state assignment in the sphere of scientific activities. The authors would like to thank V. Solovyev, A. Kayumova, I. Kayumov, P. Ivanshin, E. Utkina, and M. Matvejchuk, who have contributed to the ontology, as well as the students at Kazan Federal University, who took part in the experiments. The authors are also very grateful to A. Elizarov for his support, E. Khakimova (University of Virginia) for the assistance in writing the related work section, and three anonymous reviewers for their valuable suggestions.

\end{document}